\documentclass[10pt,twocolumn,letterpaper]{article}

\usepackage{iccv}              

\definecolor{iccvblue}{rgb}{0.21,0.49,0.74}
\usepackage[pagebackref,breaklinks,colorlinks,allcolors=iccvblue]{hyperref}

\usepackage{multicol, multirow, pifont, graphicx}
\usepackage{xcolor}
\usepackage{arydshln}
\usepackage[accsupp]{axessibility}  

\def\paperID{7} 
\def\confName{ICCV}
\def\confYear{2025}

\title{NegFaceDiff: The Power of Negative Context in Identity-Conditioned Diffusion for Synthetic Face Generation}

\author{\parbox{16cm}{\centering
    {\large Eduarda Caldeira$^1$, Naser Damer$^{1,2}$ and Fadi Boutros$^1$}\\
    {\normalsize
    $^{1}$Fraunhofer IGD, Germany, $^{2}$TU Darmstadt, Germany\\
   }    
}}

\begin{document}
\maketitle

\renewcommand\thefootnote{} 
\addtocounter{footnote}{-1}
\renewcommand\thefootnote{\arabic{footnote}}

\begin{abstract}
The use of synthetic data as an alternative to authentic datasets in face recognition (FR) development has gained significant attention, addressing privacy, ethical, and practical concerns associated with collecting and using authentic data.
Recent state-of-the-art approaches have proposed identity-conditioned diffusion models to generate identity-consistent face images, facilitating their use in training FR models. However, these methods often lack explicit sampling mechanisms to enforce inter-class separability, leading to identity overlap in the generated data and, consequently, suboptimal FR performance.
In this work, we introduce NegFaceDiff, a novel sampling method that incorporates negative conditions into the identity-conditioned diffusion process. NegFaceDiff enhances identity separation by leveraging negative conditions that explicitly guide the model away from unwanted features while preserving intra-class consistency.
Extensive experiments demonstrate that NegFaceDiff significantly improves the identity consistency and separability of data generated by identity-conditioned diffusion models. Specifically, identity separability, measured by the Fisher Discriminant Ratio (FDR), increases from 2.427 to 5.687. These improvements are reflected in FR systems trained on the NegFaceDiff dataset, which outperform models trained on data generated without negative conditions across multiple benchmarks.
\url{https://github.com/EduardaCaldeira/NegFaceDiff}.
\end{abstract}    
\section{Introduction}
\label{sec:intro}
Face recognition (FR) systems have become integral to various applications, including on-device security authentication \cite{prakash2021biometric} and automated border control (ABC-gates) \cite{DBLP:journals/cviu/GuoZ19,abs_gate}.
Traditionally, state-of-the-art (SOTA) FR models \cite{Deng_2022,DBLP:conf/cvpr/Kim0L22,wang2018cosfacelargemargincosine} have been trained on authentic face datasets collected from the internet. However, this practice raises serious privacy, data protection, and ethical concerns, as many of these datasets contain personal data obtained without explicit consent \cite{gdpr, lirias3838501}.
Several widely used authentic datasets, such as MS-Celeb-1M \cite{guo2016ms} and VGGFace2 \cite{DBLP:conf/fgr/CaoSXPZ18}, have been retracted by their creators to avoid legal and ethical consequences.
As a result, the demand for privacy-preserving and legally compliant alternatives to authentic face data has become increasingly evident.
To address this issue, synthetic face datasets \cite{DBLP:journals/inffus/MelziTVKRLDMFOZZYZWLTKZDBVGFFMUG24,DBLP:conf/fgr/Otroshi-Shahreza24,DEANDRESTAME2025103099} generated by deep generative models (DGMs) \cite{Rombach2021,FFHQ,Karras2020StyleGANADA} have emerged as a promising solution \cite{SyntheticFRSurvay}. These synthetic datasets offer several advantages, including legal compliance, and cost-effective data collection \cite{SyntheticFRSurvay,lirias3838501}.

Recent advancements in Generative Adversarial Networks (GANs) \cite{FFHQ,Karras2020StyleGANADA} and Diffusion Models (DMs) \cite{Rombach2021} have significantly enhanced the quality and diversity of synthetic face images, making them viable for training high-performance FR models. Most synthetic-based FR research has focused on generating identity-labeled face images \cite{DBLP:conf/cvpr/Kim00023,Boutros2022SFace,DBLP:conf/iccv/QiuYG00T21}, enabling supervised FR training with multi-class classification losses, such as margin penalty softmax loss and its variants \cite{Deng_2022,ElasticFace,wang2018cosfacelargemargincosine,DBLP:conf/cvpr/Kim0L22,DBLP:conf/cvpr/LiuZLL19}.
Early works, including SFace \cite{Boutros2022SFace}, SynFace \cite{DBLP:conf/iccv/QiuYG00T21}, and IDNet \cite{DBLP:conf/cvpr/KolfREBKD23}, demonstrated the feasibility of training FR models using GAN-generated face images. More recently, diffusion-based approaches such as ID$^3$ \cite{DBLP:conf/nips/Xu0WXDJHM0DH24}, IDiff-Face \cite{DBLP:conf/iccv/BoutrosGKD23}, and DCFace \cite{DBLP:conf/cvpr/Kim00023} have outperformed GAN-based methods in maintaining identity consistency and generating highly diverse identities, leading to superior FR verification performance \cite{DBLP:conf/fgr/Otroshi-Shahreza24,DBLP:journals/inffus/MelziTVKRLDMFOZZYZWLTKZDBVGFFMUG24}.
However, a key challenge remains: ensuring that synthetic images maintain high intra-class consistency while maximizing inter-class separability. Recent SOTA identity-conditioned DMs rely on positive identity conditions \cite{DBLP:conf/nips/Xu0WXDJHM0DH24,DBLP:conf/iccv/BoutrosGKD23,DBLP:conf/cvpr/Kim00023,arc2face}, encouraging the model to generate images that align with a given identity label. However, this alone does not prevent the model from generating images with unwanted features, which may lead to suboptimal inter-class separability.

In this work, we introduce NegFaceDiff, a novel sampling mechanism for identity-conditioned DMs that integrates negative contexts to enhance identity separability and thus eliminate undesired features in generated images. Inspired by the success of negative prompting \cite{ban2024understanding,koulischer2024dynamic} in text-to-image generation such as  Stable Diffusion \cite{Rombach2021} and DALLE \cite{DBLP:conf/icml/RameshPGGVRCS21}, our approach incorporates a negative context during the sampling phase to improve identity distinction. 
Similar to negative prompting in text-to-image generation \cite{ban2024understanding, DBLP:conf/icml/Wang0HG24}, we modify the denoising process to steer the generated sample away from unwanted characteristics. This is achieved by subtracting the influence of a negative condition from the latent representation during sampling. As a result, the generated images remain aligned with the desired identity (positive) while explicitly diverging from a different identity (negative).
Through extensive evaluations using various negative context selection techniques, we empirically demonstrate that incorporating negative conditions enhances both intra-class consistency and identity separability compared to using only positive identity conditions \cite{DBLP:conf/iccv/BoutrosGKD23}.
For instance, integrating negative conditions in the generation process led to a reduction in the Equal Error Rate (EER) from 0.130 to 0.039. Additionally, the mean genuine score increased from 0.226 to 0.361, while the Fisher Discriminant Ratio (FDR), which quantifies the separability between genuine and impostor scores, improved significantly from 2.427 to 5.687.
Furthermore, FR models trained on our NegFaceDiff dataset consistently achieved superior verification performance across all evaluated benchmarks compared to those trained using only positive conditions.
\section{Related Work}
\label{sec:sota}
The remarkable performance of recent DGMs \cite{Rombach2021,FFHQ,Karras2020StyleGANADA} in generating high-quality, diverse, and realistic images has driven extensive research into leveraging synthetic face images as FR training data \cite{DBLP:conf/iccv/BoutrosGKD23,DEANDRESTAME2025103099,DBLP:conf/fgr/Otroshi-Shahreza24}.
Most recent studies in this domain focus on generating identity-labeled face images using either GANs \cite{DBLP:conf/iccv/QiuYG00T21,Boutros2022SFace,ExFaceGAN} or DMs \cite{DBLP:conf/cvpr/Kim00023,DBLP:conf/iccv/BoutrosGKD23,DBLP:conf/nips/Xu0WXDJHM0DH24} to train FR in a supervised manner.
One of the earliest works proposing synthetic-based FR, SynFace \cite{DBLP:conf/iccv/QiuYG00T21},  employed DiscoFaceGAN \cite{DBLP:conf/cvpr/DengYCWT20} to generate synthetic face images and introduced identity mix-up to enhance intra-class diversity. Similarly, USynthFace \cite{DBLP:conf/fgr/BoutrosKFKD23} proposed an unsupervised approach, using extensive augmentation on a single synthetic instance to create intra-identity variations.
Other GAN-based approaches, including IDNet \cite{DBLP:conf/cvpr/KolfREBKD23}, SFace \cite{Boutros2022SFace}, and SFace2 \cite{10454585}, trained StyleGAN \cite{Karras2020StyleGANADA} in a class-conditional setting to generate identity-specific synthetic images, incorporating also knowledge distillation into their learning paradigms.
ExFaceGAN \cite{ExFaceGAN} presented a framework for disentangling identity information in the GAN latent space to generate multiple synthetic face images per identity.

Beyond GAN-based approaches, DMs have also been explored. IDiff-Face \cite{DBLP:conf/iccv/BoutrosGKD23}, DCFace \cite{DBLP:conf/cvpr/Kim00023},  GANDiffFace \cite{DBLP:conf/iccvw/MelziRTVLDS23}, ID$^3$ \cite{DBLP:conf/nips/Xu0WXDJHM0DH24} and Arc2Face \cite{arc2face} utilize DM for generating synthetic images of synthetic identities. IDiff-Face \cite{DBLP:conf/iccv/BoutrosGKD23} proposes to condition the DM on the identity context extracted from a pretrained FR. It also introduces controlled noise in the identity condition to enhance variation. DCFace \cite{DBLP:conf/cvpr/Kim00023} incorporates dual conditioning (identity and style information from real images) to control intra-class diversity. GANDiffFace \cite{DBLP:conf/iccvw/MelziRTVLDS23} integrates GAN-based image generation with diffusion-based. Arc2Face \cite{arc2face} fine-tunes a pretrained stable diffusion on the large-scale WebFace42M dataset \cite{DBLP:conf/cvpr/ZhuHDY0CZYLD021}  to generate identity-consistent, high-resolution face images. ID$^3$ \cite{DBLP:conf/nips/Xu0WXDJHM0DH24} extends IDiff-Face \cite{DBLP:conf/iccv/BoutrosGKD23} by introducing face attributes as an additional conditioning factor. 

Recognizing the significance of synthetic data in face recognition, multiple competitions have been organized to benchmark and advance synthetic FR approaches \cite{DBLP:journals/inffus/MelziTVKRLDMFOZZYZWLTKZDBVGFFMUG24,DBLP:conf/fgr/Otroshi-Shahreza24,DEANDRESTAME2025103099}. The winning solutions in these competitions utilized diffusion-based synthetic datasets \cite{DBLP:conf/iccv/BoutrosGKD23,DBLP:conf/cvpr/Kim00023} to train FR models. Despite their higher inference latency compared to GAN-based methods, DMs have demonstrated superior performance in preserving identity and producing images with high inter-class diversity.
Inspired by the success of negative prompts in text-to-image DMs \cite{ban2024understanding, DBLP:conf/icml/Wang0HG24}, we propose exploring the potential of negative contexts in identity-conditioned DMs for synthetic face generation.

\section{Methodology}
\label{sec:methodology}
This section describes NegFaceDiff, a DM sampling approach designed to generate identity-consistent images with high inter-class separability. We introduce negative conditions to the inference phase of identity-conditioned DMs, providing them with a dual guidance that simultaneously directs the generation process toward the desired identity and far away from a different identity.

\begin{figure}[ht!]
    \centering
    \includegraphics[width=0.99\linewidth]{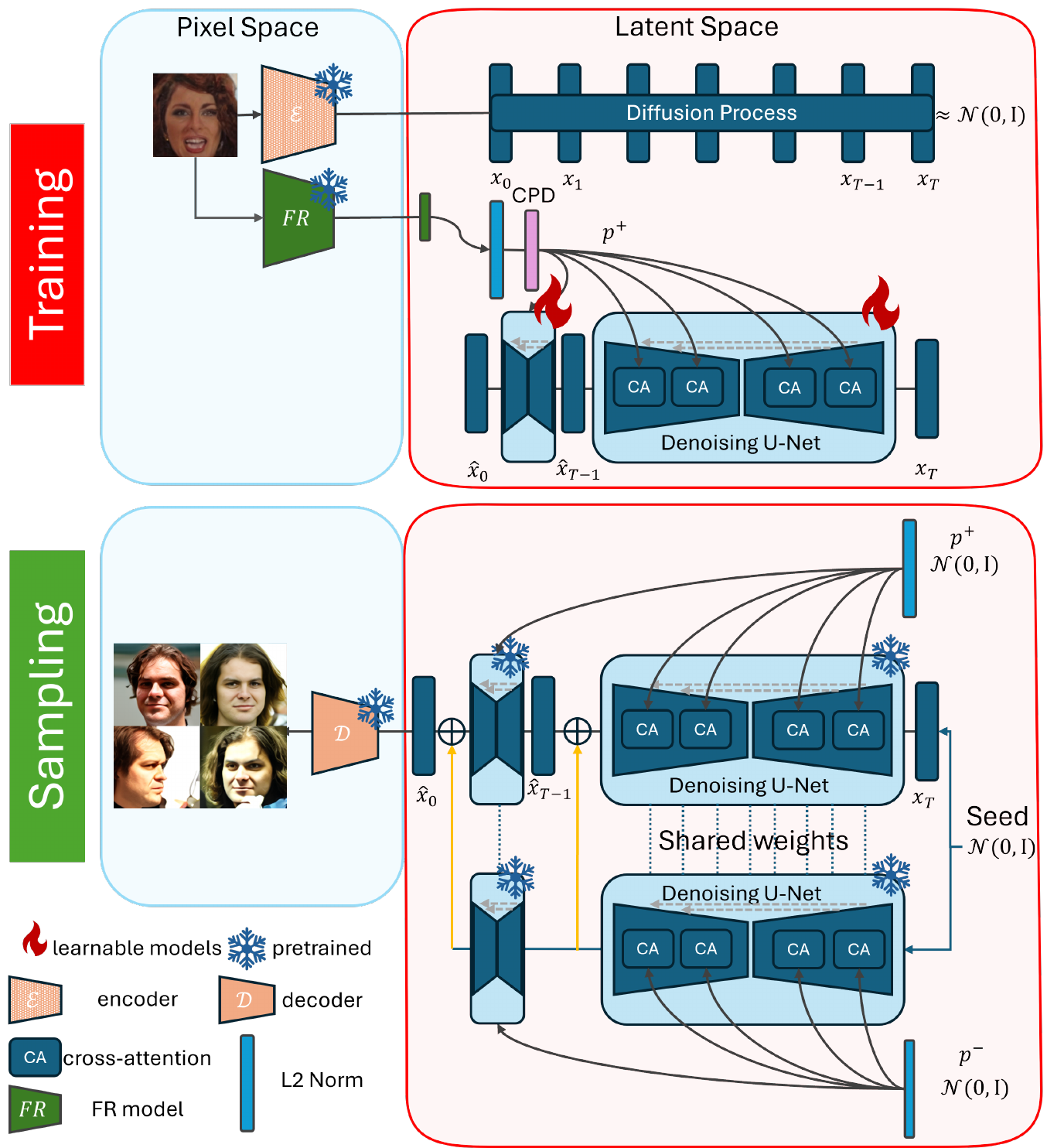}
    \caption{Overview of the proposed NegFaceDiff. \textbf{Top}: The identity-conditioned LDM \cite{DBLP:conf/iccv/BoutrosGKD23}. NegFaceDiff is a sampling approach, and thus, the training process is merely represented for illustrative purposes. \textbf{Bottom}: NegFaceDiff sampling process utilizes dual guidance of positive ($p^+$) and negative ($p^-$)  contexts. The influence of the negative context is subtracted from the latent representation (Equation \ref{eq:neg_eps}) to generate samples (Equation \ref{eq:neg_gen}).
    $p^-$  and $p^+$ are processed by the same denoising U-Net. 
    }
    \label{fig:overview}
    \vspace{-5mm}
\end{figure}

\subsection{Preliminarily: Diffusion Models}
DMs generate data through a diffusion process that gradually converts an input $x_0$ into noise (forward process) and then learns to gradually decode this noise to an approximation of the original input, $\hat{x}_0$ (reverse process). Despite being designed to generate images, DMs can be made more efficient by operating in a lower-dimensional latent space rather than directly in the pixel space, reducing the computational cost of the diffusion process and enabling the use of the same pre-trained Encoder-Decoder for several generation tasks \cite{Rombach2021}. Networks that learn the denoising process in a pre-trained autoencoder's latent space are named Latent Diffusion Models (LDM) \cite{Rombach2021}. LDMs learn how to generate a feature vector $\hat{x}_0$ that the pre-trained decoder can convert to the final image, as depicted in Figure \ref{fig:overview}.


LDMs are based on Denoising Diffusion Probabilistic Models (DDPMs) \cite{Ho2020}, which discretize the diffusion process through a finite number of steps, $T$, and learn to denoise the resultant vector $x_T$ by estimating the noise that has been added to each $x_t$ at time step $t$. LDMs' forward process (Figure \ref{fig:overview}) takes the embedding $x_0$ as input and gradually injects it with Gaussian noise through a Markov chain. This chain makes the generated embeddings $x_t, \;t>1$ converge to standard Gaussian noise assuming an infinite number of steps is performed. This effect is approximated by the LDM forward process by considering that after $T$ steps the generated embedding has approximately converged to this point ($x_T \sim \mathcal{N}$). The reverse process (Figure \ref{fig:overview}) takes $x_T$ as input and gradually learns to denoise it until $\hat{x}_0$ has been generated. 

When DMs are used to generate datasets needed to train deep neural networks under multiclass classification problems such as FR, 
it is essential to ensure that several samples belonging to the same identity can be generated and grouped into the same class. This requires some level of control over the generation process which can be achieved by conditioning it with identity information \cite{DBLP:conf/iccv/BoutrosGKD23,DBLP:conf/cvpr/Kim00023,DBLP:conf/nips/Xu0WXDJHM0DH24}. In this scenario, the identity context $p^+$ conditions the DM to generate samples of its correspondent identity (i.e., class label). To ensure that the information contained in $p^+$ is correlated with the identity present in each original sample $x$, this condition is commonly obtained by passing $x$ through a pre-trained FR model, $p^+=f(x)$ \cite{DBLP:conf/iccv/BoutrosGKD23,DBLP:conf/cvpr/Kim00023,DBLP:conf/nips/Xu0WXDJHM0DH24}.

\subsubsection{Training Phase}
Let $0<\beta_1,\beta_2,...,\beta_T<1$ be a fixed (linear) variance schedule, $\alpha_t = 1-\beta_t$ and $\overline{\alpha}_t := \sum_{i=1}^t \alpha_i$ \cite{Ho2020}. The conditional loss variant of the previously described DDPM process \cite{Ho2020} is given by:
\vspace{-1mm}
\begin{equation}
\begin{aligned}
\small
    \mathcal{L}(\theta) &:= \mathbb{E}_{t,\mathbf{x}_t, \boldsymbol{\epsilon}} 
    \left [ \left\lVert \boldsymbol{\epsilon} - \boldsymbol{\epsilon}_\theta(\mathbf{x}_t,t,\mathbf{p^+}) \right\rVert^2_2 \right ] \\
    &= \mathbb{E}_{t,\mathbf{x}_0, \boldsymbol{\epsilon}} 
    \left [ \left\lVert \boldsymbol{\epsilon} - \boldsymbol{\epsilon}_\theta 
    \left( \sqrt{\overline{\alpha}_t}\mathbf{x}_0 + \sqrt{1-\overline{\alpha}_t} 
    \boldsymbol{\epsilon}, t,\mathbf{p^+} \right)  \right\rVert^2_2 \right],
\end{aligned}
\end{equation}
where $\epsilon \sim \mathcal{N}(\mathbf{0, I})$ and $\epsilon_\theta(\mathbf{x}_t,t,\mathbf{p^+})$ is the DM prediction of the noise added at time step $t$.

\subsubsection{Sampling Phase}
Once the model is trained, a synthetic image $x$ can be generated by sampling a random seed from a Gaussian distribution and passing it through the denoising process. A set of samples belonging to the same identity can be sampled by fixing the identity condition $p^+$ and varying the initial random seed.
This process mirrors a score-based sampling chain with \textit{Langevin dynamics} and thus includes additional noise vectors $\boldmath{\zeta}_t\sim \mathcal{N}(\mathbf{0,I})$ at each time step $t$ \cite{DBLP:conf/iccv/BoutrosGKD23}:
\vspace{-1mm}
\begin{equation}
\begin{aligned}
\small
    \mathbf{x}_{t-1} &= {\mu}_\theta(\mathbf{x}_t,t,\mathbf{p^+}) + \sigma_t {\zeta}_t \\
&= \frac{1}{\sqrt{\alpha_t}} \left ( \mathbf{x}_{t} - \frac{1-\alpha_t}{\sqrt{1-\overline{\alpha}_t}} \epsilon_\theta(\mathbf{x}_t, t, \mathbf{p^+}) \right ) + \sigma_t {\zeta}_t .
\label{eq:sampling_only_pos}
\end{aligned}
\end{equation}

\subsection{NegFaceDiff: The Power of Negative}


Negative conditions have been successfully used to provide extra guidance to the sampling process in tasks such as text-to-image generation \cite{ban2024understanding, DBLP:conf/icml/Wang0HG24}, avoiding the presence of features from the negative condition. When generating identity-specific face images, the sampling process can benefit from additional guidance (negative condition) toward the exclusion of undesired characteristics in the generated data, which the positive condition alone does not explicitly provide. As proved in Section \ref{sec:separability}, using an additional negative identity condition results in better separation between identities since the generative model is also instructed not to generate undesired features. Simultaneously, negative conditions can restrain the space occupied by each identity without eliminating intra-class variability. This leads to a higher control over identity consistency and results in better identity separation, which facilitates the FR task, as proved in Sections \ref{sec:eval_ddim} and \ref{sec:res_sota}.

In this work, we explore the inclusion of negative identity conditions as additional guidance to the positive condition, aiming at enhancing the identity-separability and consistency of identity-conditioned DMs. To that end, we propose the first sampling method that incorporates negative conditions in identity-conditioned DMs, NegFaceDiff. This is achieved by redefining $\epsilon_\theta$ to subtract the effect of the added noise prediction when $p^-$ is used as an identity condition from the added noise prediction when considering the positive condition $p^+$:
\vspace{-1mm}
\begin{equation}
\label{eq:neg_eps}
\small
    \hat{\epsilon}_\theta(x_t, t, p^+, p^-) = (1 + w) \, \epsilon_\theta(x_t,t,p^+) - w \, \epsilon_\theta(x_t,t,p^-),
\end{equation}
where $w$ represents the guidance strength of the negative condition \cite{ban2024understanding, DBLP:conf/icml/Wang0HG24}. The subtraction term ensures that the generated images follow $p^+$ and avoid features associated with $p^-$. Formally, we extend the sampling process described by Equation \ref{eq:sampling_only_pos}: 
\vspace{-1mm}
\begin{equation}
\label{eq:neg_gen}
\begin{aligned}
\small
    \mathbf{x}_{t-1} &= {\mu}_\theta(\mathbf{x}_t,t,\mathbf{p^+},\mathbf{p^-}) + \sigma_t {\zeta}_t \\
&= \frac{1}{\sqrt{\alpha_t}} \left ( \mathbf{x}_{t} - \frac{1-\alpha_t}{\sqrt{1-\overline{\alpha}_t}} \,\hat{\epsilon}_\theta(x_t, t, p^+, p^-)\right ) + \sigma_t {\zeta}_t.
\end{aligned}
\end{equation}



\subsection{Negative Context Selection}
\label{sec:neg_selection}
As our work aims to generate datasets for FR training with multiple samples belonging to $N$ synthetic identities, we uniformly sample $N$ positive identity conditions $p^+$ from the normal distribution, following \cite{DBLP:conf/iccv/BoutrosGKD23}. 
After selecting $p^+$, the negative condition $p^-$ associated with each identity can correspond to any different identity ($p^-\neq p^+$). Hence, for each $p^+$, $p^-$ can be selected from $N-1$ possible contexts.
Since $p^+$ and $p^-$ compete to impose their effect in the sampling process, we hypothesize that the distance between these two conditions plays a crucial role in the effect provided by $p^-$. 

DMs that incorporate negative conditions in their sampling process \cite{ban2024understanding, DBLP:conf/icml/Wang0HG24} often formulate them in an extreme manner by explicitly specifying the characteristics that should not be present in the generated images. Taking text-to-image generation as an example, the negative condition ``Eiffel Tower'' would be expected to completely avoid this monument in the generated sample \cite{ban2024understanding}. With this in mind, we hypothesize that NegFaceDiff will be more efficient when $p^-$ and $p^+$ correspond to identities with substantially different features so that $p^-$ highlights the characteristics that should not be present in the generated samples. Hence, we propose a negative condition selection strategy, Far-Neg, which defines $p^-$ as the farthest context for each $p^+$. We further analyze three alternative selection strategies to validate our hypothesis on Far-Neg effectiveness. Close-Neg selects $p^-$ as the closest context for each positive condition $p^+$. Rand-Neg and Mid-Neg constitute two intermediate options that are expected to balance the properties associated with Close-Neg and Far-Neg. Rand-Neg randomly selects $p^-$ for each $p^+$ while Mid-Neg defines it as the median distance impostor.
Note that Rand-Neg is an effective selection approach, as we present in our Section \ref{sec:separability}, in case the complete set of contexts is not accessible in advance or the generation process outputs images of a single identity.  
As proved in Section \ref{sec:eval_ddim}, Far-Neg results in better guidance of the sampling process toward the removal of unwanted characteristics, leading to higher identity consistency and supporting our initial conjecture. 

The selection process of $p^-$ following Close-Neg and Far-Neg is given by Equations \ref{eq:close_neg} and \ref{eq:far_neg}, respectively, where $d(.)$ represents the selected distance metric, in our case the normalized Euclidean distance:
\begin{equation}
    p^- = \arg\min_{p_j \neq p^+} d(p^+, p_j), \quad p_j \in \{p_0, \dots, p_{N-1}\}
    \label{eq:close_neg}.
\end{equation}
\vspace{-3mm}
\begin{equation}
    p^- = \arg\max_{p_j \neq p^+} d(p^+, p_j), \quad p_j \in \{p_0, \dots, p_{N-1}\}
    \label{eq:far_neg}.
\end{equation}
For Mid-Neg, the normalized Euclidean distance values of all possible pairs $(p^+, p_j)$ for a specific $p^+$ are ordered and $p^-$ is selected as the $p_j$ that results in the central value of the ordered distance list. We further analyze the results achieved when $p^-$ is defined as the null vector (Null), to assess the performance of NegDiffProb in comparison to this scenario \cite{ban2024understanding}.



\section{Experimental Setup}

\subsection{Baseline Diffusion Models}
\label{sec:setup_dms}
We used a pre-trained IDiff-Face \cite{DBLP:conf/iccv/BoutrosGKD23} with a contextual partial dropout of 25\% as the base DM. IDiff-Face is a conditional LDM trained on the latent space of a pre-trained Auto-Encoder \cite{Rombach2021} and conditioned on identity contexts, i.e., embeddings from a pre-trained FR model. This model was trained on the Flickr-Faces-HQ (FFHQ) dataset \cite{FFHQ} containing 70K images. 
In this work, we further 
trained IDiff-Face \cite{DBLP:conf/iccv/BoutrosGKD23} on CASIA-WebFace (C-WF) dataset \cite{DBLP:journals/corr/YiLLL14a}, following training setups described in \cite{DBLP:conf/iccv/BoutrosGKD23} to provide comparable results with previous works that utilized DM trained on C-WF.
IDiff-Face \cite{DBLP:conf/iccv/BoutrosGKD23} generates a novel synthetic identity by using a synthetic embedding, i.e., a random vector from a normal distribution.
To generate a set of images belonging to the same identity, IDiff-Face samples a vector from a normal distribution and uses it as an identity condition. This context is fixed for all images of the same identity.
Our choice for IDiff-Face \cite{DBLP:conf/iccv/BoutrosGKD23} is due to its simplicity, to its wide adaption in many solutions submitted to several international competitions \cite{DBLP:journals/inffus/MelziTVKRLDMFOZZYZWLTKZDBVGFFMUG24,DBLP:conf/fgr/Otroshi-Shahreza24,DEANDRESTAME2025103099} and to its SOTA performances (e.g., winner solution in SDFR competition \cite{DBLP:conf/fgr/Otroshi-Shahreza24}).   
We extended the sampling mechanism of the IDiff-Face with a negative context as described in Section \ref{sec:methodology}. 
Note that IDiff-Face utilized a DDPM \cite{Ho2020} with 1000 steps for training and sampling. In our ablation studies, we utilized a Denoising Diffusion Implicit Model (DDIM) \cite{DBLP:conf/iclr/SongME21} in the sampling phase with 200 steps, enabling the investigation of different negative sampling techniques. DDIM improves the efficiency of DDPM by introducing a non-Markovian process, allowing for deterministic and faster sampling. It enables image generation with significantly fewer steps while maintaining, to a large degree, quality. 
The SOTA results (Table \ref{tab:sota}) are reported using DDPM to provide a fair comparison with previous works.   
Using each of the sampling techniques described in Section \ref{sec:methodology}, we generate 10k novel identities, 50 images per identity.

The introduction of a negative context in the DM's sampling process also requires defining a value for $w$. 
The value of $w$ was selected by fixing one of the proposed NegFaceDiff negative context selection strategies, Far-Neg, and using it to generate datasets with distinct $w$ values ($w=\{0.25, 0.50, 1.00\}$). These datasets were then used to train FR systems, in order to assess which $w$ value results in a dataset better prepared for FR training. The evaluated settings resulted in average performances on small-scale datasets of 84.41\% ($w=0.25$), 85.22\% ($w=0.50$) and 78.79\% ($w=1.00$). Hence, we fixed $w=0.5$ for all the remaining experiments.




\begin{figure}[t!]
    \includegraphics[width=0.97\linewidth]{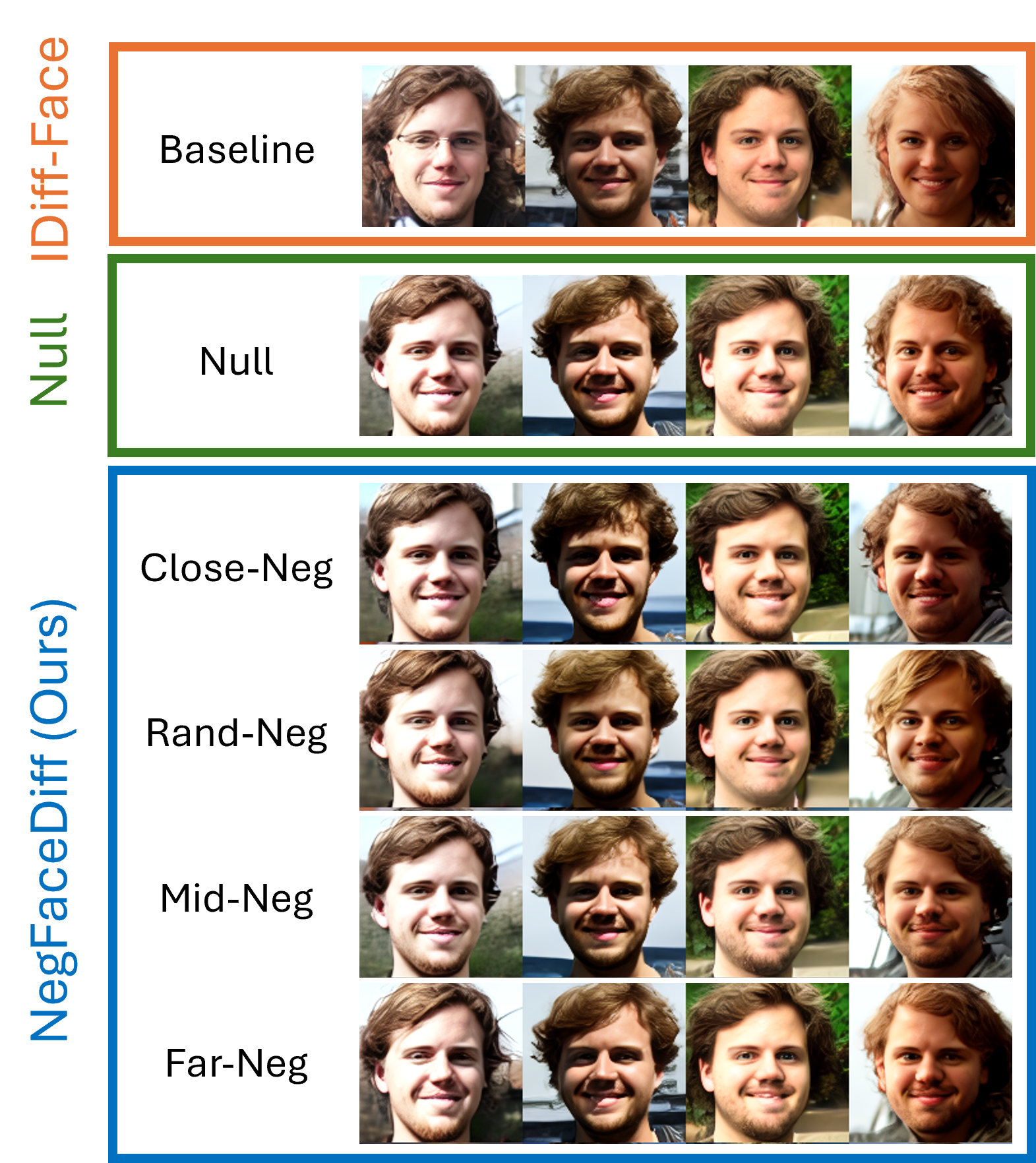}
    \vspace{-2mm}
    \caption{Samples generated by NegFaceDiff (blue box) using different negative selection methods (Close-Neg, Rand-Neg, Mid-Neg and Far-Neg). The samples in the orange box are generated by IDiff-Face, which only makes use of the positive condition $p^+$, and the samples in the green box are generated by Null, which uses the positive context $p^+$ and a null vector as  $p^-$. All the presented samples were generated with the same $p^+$ and thus represent the same identity. One can verify that NegFaceDiff is an efficient method in eliminating undesired variations, e.g., the gender swap in the fourth sample, without compromising the identity consistency of the remaining samples.}
    \label{fig:faces}
    \vspace{-3mm}
\end{figure}

\subsection{Identity-Separability Evaluation}
\label{sec:setup_idsep}
We evaluate the identity-separability of our synthetic data generated with the different negative sampling techniques and compare it to the case where data is generated without a negative condition \cite{DBLP:conf/iccv/BoutrosGKD23}.
The verification performances are reported as FMR100 and FMR1000, which are the lowest false non-match rate (FNMR) for a false match rate (FMR)$\leq$1.0\% and $\leq$0.1\%, respectively, along with the Equal Error Rate (EER). We also plot the histograms of genuine and imposter score distributions.
The inter-class separability and intra-class compactness are evaluated by reporting the genuine and imposter means (G-mean and I-mean) and standard deviation (G-STD and I-STD), respectively. Additionally, we report the FDR \cite{poh2004study} to quantify the separability of genuine and imposters scores. 
We used a pre-trained ResNet100 \cite{He2015DeepRL} with ElasticFace \cite{ElasticFace} on MS1MV2 \cite{Deng_2022,guo2016ms} to extract the feature representations needed for the evaluation described above. 

\subsection{NegFaceDiff FR Training Setup}
\label{sec:fr_eval}

All presented FR models in this paper use ResNet50 \cite{He2015DeepRL} as a network architecture and
are trained with the CosFace loss \cite{wang2018cosfacelargemargincosine}. We set the mini-batch size to 512 and train with a Stochastic Gradient Descent (SGD) optimizer, setting the momentum to 0.9 and the weight decay to 5e-4, following \cite{ExFaceGAN}. The margin penalty of CosFace loss was set to 0.35 and the scale factor to 64 \cite{wang2018cosfacelargemargincosine}, following \cite{DBLP:conf/iccv/BoutrosGKD23}. The models are trained on our synthetically generated datasets and compared to the models trained with data sampled without the negative context. 
All FR models in this paper are trained with 500K images (10k identities with 50 images per identity). 
During training we applied Random Augmentation with 4 operations and a magnitude of 16, following \cite{ExFaceGAN}, unless otherwise explicitly noted.

\subsection{Evaluation Benchmarks and Metrics}
\label{sec:eval_benchmarks}
The performance of the FR models trained in this work is evaluated and reported as the verification accuracy on five benchmarks, Labeled Faces in the Wild LFW \cite{LFWDatabase}, AgeDb-30 \cite{AgeDB30Database}, Cross-Age LFW (CA-LFW) \cite{CALFWDatabase}, Celebrities in Frontal-Profile in the Wild (CFP-FP) \cite{CFPFPDatabase}, and Cross-Pose LFW (CP-LFW) \cite{CPLFWDatabase}, following their official evaluation protocol. 
In addition, we evaluated on the large-scale evaluation benchmark IARPA Janus Benchmark–C (IJB-C) \cite{DBLP:conf/icb/MazeADKMO0NACG18}. For IJB-C, we used the official 1:1 mixed verification protocol and reported the verification performance as True Acceptance Rates (TAR) at False Acceptance Rates (FAR) of 1e-4 and 1e-5. We enrich our evaluation with demographic bias assessment on the Racial Faces in the Wild (RFW) \cite{RFW} dataset by reporting the verification accuracies on its four subsets (African, Caucasian, Indian, and Asian) as well as bias-assessing metrics, namely the standard deviation (STD) and the skewed error ratio (SER) \cite{RFW}. The STD quantitatively measures the performance variation between the four demographic groups. The SER facilitates comprehension of the relative differences in the intra-ethnicity accuracies by contrasting the best and worst classification mistakes: $SER=\frac {100-min(acc)}{100-max(acc)}$, where $acc$ is a vector containing the four intra-ethnicity accuracy values.
\section{Results and Discussion}
This section presents the results achieved in this paper. We first evaluate the identity separability of the data generated using dual conditions (our NegFaceDiff) and by the baseline model (IDiff-Face \cite{DBLP:conf/iccv/BoutrosGKD23}). This is followed by comparisons with recent SOTA works proposing synthetic-based FR. 

\subsection{Ablation Studies}
We first investigate the impact of dual conditions, negative and positive, on generating identity-separable images using identity-conditioned DMs. We compare this to the case where only the positive condition is applied (IDiff-Face \cite{DBLP:conf/iccv/BoutrosGKD23}). We begin by analyzing the influence of the different $p^-$ selection methods described in Section \ref{sec:neg_selection} (Close-Neg, Rand-Neg, Mid-Neg, and Far-Neg) on identity separability. Subsequently, the datasets generated by each approach are used to train a FR model. All the ablation study results were obtained by sampling from a DM trained on FFHQ.

\tabcolsep=0.1cm
\begin{table}[t!]
\centering
\resizebox{\linewidth}{!}{%
    \begin{tabular}{@{}lccccccccccc@{}}\toprule
    \multicolumn{1}{c}{\textbf{ }}& \phantom{abc} & \multicolumn{3}{c}{\textbf{Operation Metrics}} & \phantom{abc} & \phantom{abc} & \multicolumn{5}{c}{\textbf{Score Distributions}}\\
    \cmidrule{7-12}
    \multicolumn{5}{c}{\textbf{ }}& \phantom{abc} & \multicolumn{2}{c}{genuine} &   \multicolumn{2}{c}{imposter} &  \multicolumn{1}{c}{\textbf{ }}\\
    
    \cmidrule{1-1} \cmidrule{3-5} \cmidrule{7-12} 
    \textbf{Method} && EER $\downarrow$  & FMR100 $\downarrow$  & FMR1000 $\downarrow$  && mean & std & mean & std & FDR $\uparrow$ \\
    \midrule
 C-WF \cite{DBLP:journals/corr/YiLLL14a} && 0.076  & 0.092  & 0.107   &&  0.536 &  0.215  &  0.003  &  0.070  &  5.541 \\

    LFW \cite{LFWDatabase} &&  0.002   &  0.002  &  0.002  &&  0.708  &  0.099  &  0.003  &  0.070  &  33.301 \\

    \cmidrule{1-1} \cmidrule{3-5} \cmidrule{7-12}
    IDiff-Face \cite{DBLP:conf/iccv/BoutrosGKD23} &&  0.130   &  0.385  &  0.607  &&  0.226  &  0.117 &  0.014  &  0.070  &  2.427 \\
    \cmidrule{1-1} \cmidrule{3-5} \cmidrule{7-12}
    Null && \textbf{0.034} & \textbf{0.066} & \textbf{0.158} && 0.377 & 0.127 & 0.023 & 0.060 & \textbf{6.361} \\
    \cmidrule{1-1} \cmidrule{3-5} \cmidrule{7-12}
    \textbf{NegFaceDiff} &&&&&&&&&&& \\
    \midrule
    Close-Neg && 0.043 & 0.094 & 0.208 && 0.345 & 0.129 & 0.020 & 0.059 &  5.280 \\
    Rand-Neg && 0.040 & 0.085 & 0.194 && 0.354 &  0.129  & 0.020 & 0.059 &  5.513 \\
    Mid-Neg && 0.041 & 0.083 & 0.186 &&  0.355 & 0.130  & 0.019 & 0.059 &  5.526 \\
    Far-Neg && \underline{0.039} & \underline{0.078} & \underline{0.179} && 0.361 & 0.131 & 0.019 & 0.059 &  \underline{5.687}  \\
    \bottomrule
    \end{tabular}
}
\caption{Evaluation of the identity-separability of datasets generated by the different negative condition selection methods of NegFaceDiff. The first two rows present the results on authentic LFW \cite{LFWDatabase} and C-WF \cite{DBLP:journals/corr/YiLLL14a} and are provided as a reference. The third row provides the evaluation results of IDiff-Face \cite{DBLP:conf/iccv/BoutrosGKD23}, which is considered as a baseline, as it does not use negative conditions. All the synthetic evaluations are based on a synthetically generated dataset with $1{,}000$ identities and $20$ sampled images per identity. The lowest errors and the highest genuine-imposter separability scores (FDR) on synthetic datasets are marked in \textbf{bold}. The second best per column is \underline{underlined}. One can clearly notice that NegFaceDiff achieved lower EER and higher FDR in comparison to the baseline (IDiff-Face \cite{DBLP:conf/iccv/BoutrosGKD23}). The highest identity separability was achieved with Null.}
\label{tab:id_sep}
\vspace{-3mm}
\end{table}


\paragraph{Identity Separability}
\label{sec:separability}

\begin{figure*}
    \includegraphics[width=0.95\linewidth]{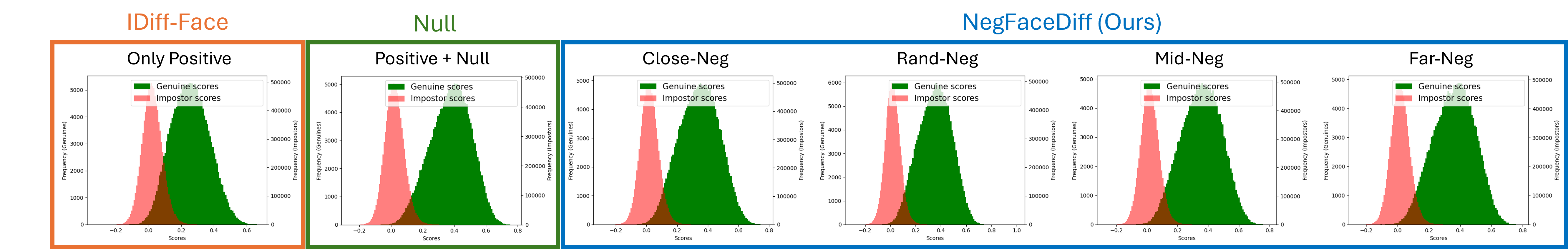}
    \vspace{-2mm}
    \caption{Histograms of the genuine and impostor score distributions for IDiff-Face \cite{DBLP:conf/iccv/BoutrosGKD23} (baseline), Null, and the  NegFaceDiff (Close-Neg, Rand-Neg, Mid-Neg, and Far-Neg). One can notice the improvement in the identity-separability by data generated using the dual condition (positive and negative), in comparison to the data generated using only the positive condition (IDiff-Face \cite{DBLP:conf/iccv/BoutrosGKD23}), with the smallest overlap between genuine and impostor distributions achieved by Null.    
    These results are visual representations of the ones reported in Table \ref{tab:id_sep}.}
    \label{fig:dist}
    \vspace{-3mm}
\end{figure*}

Figure \ref{fig:dist} shows the histograms of genuine and imposter score distributions of data generated using positive conditions (IDiff-Face \cite{DBLP:conf/iccv/BoutrosGKD23}), a combination of positive and null contexts (Null) and NegFaceDiff.
These visual results are numerically supported in Table \ref{tab:id_sep} by the verification performance metrics described in Section \ref{sec:setup_idsep}

As shown in Figure \ref{fig:dist}, NegFaceDiff methods yield better separability between genuine and impostor score distributions when compared to IDiff-Face \cite{DBLP:conf/iccv/BoutrosGKD23}, resulting in a smaller overlap between the two score distributions. Compared to IDiff-Face \cite{DBLP:conf/iccv/BoutrosGKD23}, the genuine distributions of NegFaceDiff shift towards higher values, which is quantitatively supported by their higher G-Mean values, with the highest value being achieved by Far-Neg, as presented in Table \ref{tab:id_sep}.
This trend is further confirmed by the significant differences observed between IDiff-Face \cite{DBLP:conf/iccv/BoutrosGKD23} and NegFaceDiff in the evaluation metrics shown in Table \ref{tab:id_sep}. Furthermore, a comparison of the four proposed negative selection methods for NegFaceDiff reveals that Far-Neg consistently outperforms the others, suggesting that enforcing the exclusion of features inconsistent with the desired identity enhances identity consistency while improving inter-class separability. It can also be observed that Null results in the smallest overlap between genuine and impostor distributions (Figure \ref{fig:dist}) and consistently surpasses the remaining methods in the metrics highlighted in Table \ref{tab:id_sep}. Hence, this method results in the highest inter-class separability, reducing the EER achieved by IDiff-Face \cite{DBLP:conf/iccv/BoutrosGKD23} to approximately 26\% and increasing the FDR by 2.6 times. Notably, the second best-performing method, NegFaceDiff (Far-Neg), reduced the EER to approximately 30\% of the value achieved without negative conditions, while increasing the FDR by more than 2.3 times. Furthermore, NegFaceDiff (Far-Neg) led to a higher genuine standard deviation than the remaining techniques, suggesting a higher intra-identity variation of the generated data, which, together with high inter-class separability, is desired in FR training datasets, as empirically proved in the next ablation study.

Samples of the same identity generated by IDiff-Face \cite{DBLP:conf/iccv/BoutrosGKD23}, Null and our NegFaceDiff are shown in Figure \ref{fig:faces}. One can notice that the represented identity is more consistent when using dual conditions, negative and positive, compared to data generated with only positive conditions. 


\tabcolsep=0.04cm
\begin{table}\centering
\resizebox{\linewidth}{!}{\begin{tabular}{@{}ccclccccccccccc@{}}
\cmidrule{6-14} 
\multicolumn{2}{c}{}& \phantom{abc} & && \multicolumn{9}{c}{\textbf{Verification Benchmarks $\uparrow$}}\\

\cmidrule{6-14}
\multicolumn{4}{c}{} & \phantom{abc} & \phantom{abc} & \phantom{abc} &\multicolumn{2}{c}{Cross-Age}  & \phantom{abc} & \multicolumn{2}{c}{Cross-Pose}  & \phantom{abc} & \phantom{abc}\\

 \cmidrule{8-9}  \cmidrule{11-12} 
 \textbf{Diffusion} &\textbf{Aug.} & \multicolumn{2}{c}{\textbf{Sampling Method}} && LFW && AgeDB-30  & CA-FLW && CFP-FP  & CP-LFW  && Average\\
\midrule
\multirow{14}{*}{DDIM} & \multirow{7}{*}{\ding{56}} & \multicolumn{2}{c}{IDiff-Face \cite{DBLP:conf/iccv/BoutrosGKD23}} &&  96.30	 && 78.15    &  86.23 && 81.51 & 77.57     && 83.95 \\
\cmidrule{3-4} \cmidrule{6-6} \cmidrule{8-9}  \cmidrule{11-12} \cmidrule{14-14} 
& & \multicolumn{2}{c}{Null} && \underline{96.77} && 78.80 & 86.65 && 81.01 & 77.10 &&	84.07\\
\cmidrule{3-4} \cmidrule{6-6} \cmidrule{8-9}  \cmidrule{11-12} \cmidrule{14-14} 
& & \textbf{NegFaceDiff} & Close-Neg &&  96.63 && \underline{80.23} &  87.22 && \underline{82.91} &  78.28 && \underline{85.06} \\
& & {} & Rand-Neg && 96.47  &&  79.88   &  \underline{87.45} && 82.40 & \underline{78.30} && 84.90 \\
& &{} & Mid-Neg && 96.67 &&   79.28  & 86.67  && \textbf{83.36} &  78.18    && 84.83\\
& &{} & Far-Neg  && \textbf{96.78}  && \textbf{81.35}  & \textbf{87.67}  && 81.89 &  \textbf{78.43} && \textbf{85.22}\\

\cmidrule{2-14}

 & \multirow{7}{*}{\ding{52}} & \multicolumn{2}{c}{IDiff-Face \cite{DBLP:conf/iccv/BoutrosGKD23}} && 97.65 && 86.53 & 90.13 && \textbf{84.06} & 78.98 && 87.47 \\
\cmidrule{3-4} \cmidrule{6-6} \cmidrule{8-9}  \cmidrule{11-12} \cmidrule{14-14} 
& & \multicolumn{2}{c}{Null} && \underline{97.68} && 86.10 & 90.18 && 83.46 & 79.27 &&	87.34\\
\cmidrule{3-4} \cmidrule{6-6} \cmidrule{8-9}  \cmidrule{11-12} \cmidrule{14-14} 
& & \textbf{NegFaceDiff} & Close-Neg && 97.07 && 86.28 & 90.15 && 83.69 & 78.42 && 87.25 \\
& & {} & Rand-Neg && \textbf{97.72}	  &&   \textbf{86.92}		  & \underline{90.25}	  && 83.64  &  79.42 && \underline{87.59} \\
& &{} & Mid-Neg && 	97.45  &&  \underline{86.62}  & 89.83 && 83.59 & \underline{80.12}  &&  87.52 \\
& &{} & Far-Neg  &&  97.60	 && 86.45 & 		\textbf{90.27}  &&  \underline{83.73}	   &    \textbf{80.20}	  && \textbf{87.65}\\

\hline \hline
\multirow{2.5}{*}{DDPM} & \ding{56} & \multirow{2.5}{*}
{\textbf{NegFaceDiff}} & \multirow{2.5}{*}{Far-Neg} && 96.87 && 79.95 & 87.65 && 83.00 & 78.50 && 85.19\\
\cmidrule{5-14}
& \ding{52} & {} &&& 97.97 &&  86.87 & 90.77  && 85.83 & 81.07 && 88.50 \\
\bottomrule
\end{tabular}}
\caption{Verification accuracies (in $\%$) on five FR benchmarks for models trained on 500k samples (10k identities, 50 images per identity) generated with different negative condition selection strategies. IDiff-Face \cite{DBLP:conf/iccv/BoutrosGKD23} results are presented as a baseline that does not include negative conditions in the generation. The best accuracies are marked in \textbf{bold} and the second best are \underline{underlined}.}
\label{tab:ddim_comparison}
\vspace{-3mm}
\end{table}

\paragraph{Face Recognition}
\label{sec:eval_ddim}
We evaluate the proposed NegFaceDiff variants in terms of their effectiveness in generating synthetic training datasets for FR. To this end, the datasets generated with dual conditions (Close-Neg, Rand-Neg, Mid-Neg, and Far-Neg), by Null and by IDiff-Face \cite{DBLP:conf/iccv/BoutrosGKD23} are used to train FR models, whose verification performances are reported in Table \ref{tab:ddim_comparison}. Given that DMs are relatively slow at sampling \cite{DBLP:conf/iclr/SongME21} and to enable deep investigations, we generate synthetic datasets using 200 steps of DDIM \cite{DBLP:conf/iclr/SongME21}, which requires $\sim$5 times less inference time than the 1000 steps of DDPM \cite{Ho2020}. The samples of the best-performing setting (Far-Neg) were later regenerated using 1000 steps DDPM.

When data is generated using DDIM with 200 steps, FR models trained without data augmentation on the NegFaceDiff outperformed the model trained on data generated by IDiff-Face \cite{DBLP:conf/iccv/BoutrosGKD23}. NegFaceDiff also surpassed the second baseline, Null, which uses a null condition as negative to perform the sampling, showing that NegFaceDiff provides a better trade-off between inter-class separability and intra-class variability than NU, as previously discussed in this section. The average verification accuracy increased from 83.95\% (IDiff-Face \cite{DBLP:conf/iccv/BoutrosGKD23}) and 84.07\% (Null) to 85.22\% using our NegFaceDiff (Far-Neg), as presented in Table \ref{tab:ddim_comparison}. These results highlight the benefits of incorporating negative conditional information in the sampling process. It can also be seen that augmenting the training datasets enhanced the verification accuracies, confirming the results reported in \cite{DBLP:conf/iccv/BoutrosGKD23}.

The best overall performance was achieved by NegFaceDiff (Far-Neg), confirming its effectiveness in producing synthetic datasets well-suited for FR training. Thus, we chose Far-Neg as a negative condition selection method and only refer to the results of NegFaceDiff with Far-Neg selection from this point onward. Nonetheless, it is important to highlight that Rand-Neg presents the advantage of not requiring knowledge regarding the set of $p^+$ used for sampling in advance. This strategy constitutes the second best-performing selection method and can thus be effectively used in scenarios where the complete set of contexts $p^+$ is not accessible in advance or when the generation process outputs images of a single identity. Given the best-performing model (Far-Neg), we regenerate the data using DDPM and the exact same contexts. The model trained on data generated by DDPM achieved very competitive results compared to the model trained on data generated by DDIM when no data augmentation for FR training is applied (Table \ref{tab:ddim_comparison}). This performance is slightly improved when data augmentation is used for FR training.

\begin{table*}[ht!]
\centering
\resizebox{0.99\textwidth}{!}{%
\footnotesize
\begin{tabular}{cccccccccc|cc}
\hline
    \multirow{2}{*}{\textbf{Data Generation}} & 
    \multirow{2}{*}{\textbf{Method}} &
  \multirow{2}{*}{\textbf{DGMs Train Dataset}} &
  \multirow{2}{*}{\textbf{Id/Img.}} &
  \multirow{2}{*}{\textbf{LFW}} &
  \multirow{2}{*}{\textbf{AgeDB}} &
  \multirow{2}{*}{\textbf{CFP-FP}} &
  \multirow{2}{*}{\textbf{CA-LFW}} &
  \multirow{2}{*}{\textbf{CP-LFW}} &
  \multirow{2}{*}{\textbf{Avg}} &
  \multicolumn{2}{c}{\textbf{IJB-C}} \\
  &&&&&&&&&&
  \textbf{$10^{-5}$} &
  \textbf{$10^{-4}$} \\ \hline
 Authentic &
 C-WF \cite{DBLP:journals/corr/YiLLL14a} &
  - &
  10.5K/46 &
  99.55 &
  94.55 &
  95.31 &
  93.78 &
  89.95 &
  94.63 & 93.96 & 96.05
  \\ \hline
 \multirow{2}{*}{Digital Rendering}                & DigiFace-1M \cite{DigiFace1M}    & - &10K/50   & 95.40 & 76.97 & 87.40 & 78.62 & 78.87 & 83.45 & - & - \\ 
  &    DigiFace-1M \cite{DigiFace1M}*              & - & 10K/50   & 91.15 & 74.00 & 82.93 & 75.30 & 73.40 & 79.36 & 30.01 &  44.78 \\ \hline
\multirow{7}{*}{GAN} &   SynFace \cite{DBLP:conf/iccv/QiuYG00T21}        & FFHQ &10K/50   & 88.98 & -    & -     & -     & -     & -  & - & -   \\
  &                 SynFace (w/IM) \cite{DBLP:conf/iccv/QiuYG00T21}                 & FFHQ &10K/50   & 91.93 & 61.63 & 75.03 & 74.73 & 70.43 & 74.75 & - & - \\
   &               USynthFace \cite{DBLP:conf/fgr/BoutrosKFKD23}                   & FFHQ & 400K/1   & 92.23 & 71.62 & 78.56  & 77.05 & 72.03 & 78.30 & - & - \\
      &             ExFaceGAN(Con)       \cite{ExFaceGAN}                         & FFHQ & 10K/50   & 93.50  & 78.92  & 73.84 & 82.98  & 71.60 & 80.17 & 12.92 & 43.28 \\  \cdashline{2-12}
  & IDnet \cite{DBLP:conf/cvpr/KolfREBKD23}    &               C-WF &10.5K/50 & 92.58 & 73.53 & 75.40  & 79.90 & 74.25 & 79.13 & 38.85 & 53.25 \\
            &              SFace \cite{Boutros2022SFace}                     &  C-WF & 10.5K/60 & 91.87 & 71.68 & 73.86 & 77.93 & 73.20  & 77.71 & 12.70 & 19.87\\
                      &             SFace2+ \cite{10454585}                       & C-WF &10.5K/60 & 95.60  & 77.37 &  77.11  & 83.40 & 74.60 & 81.62 & 0.85 & 5.36\\ \hline
 \multirow{8}{*}{DM}  & Arc2Face \cite{arc2face} &  WebFace4M + FFHQ + CelebA & 10K/50 & 98.81 & 90.18 & 91.87 & 92.63 & 85.16 & 91.73 & - & - \\ \cdashline{2-12}
 &            ID$^3$ \cite{DBLP:conf/nips/Xu0WXDJHM0DH24}                      & FFHQ & 10K/50   & 97.28 & 83.78 & 85.00 & 89.30 & 77.13 & 86.50 & - & - \\
& IDiff-Face  \cite{DBLP:conf/iccv/BoutrosGKD23}   & FFHQ & 10K/50   & \textbf{98.00} & 86.43 & \textbf{85.47} & \textbf{90.65} & 80.45 & \textbf{88.20} & 20.60 & 62.60 \\  
  &            \textbf{NegFaceDiff (Ours)}                          & FFHQ & 10K/50 & 97.60 &	\textbf{86.53} &	85.33 &	90.28 &	\textbf{80.73} &	88.10 & \textbf{58.09} & \textbf{73.93} \\ \cdashline{2-12}
&   DCFace  \cite{DBLP:conf/cvpr/Kim00023}     & C-WF &10K/50   & 98.55 & 89.70 & 85.33 & 91.60 & 82.62 & 89.56 & 60.80* & 74.63* \\ 
&                 ID$^3$ \cite{DBLP:conf/nips/Xu0WXDJHM0DH24}                  & C-WF & 10K/50 & 97.68 & \textbf{91.00} & 86.84 & 90.73 & 82.77 & 89.80 & - & - \\
  &   IDiff-Face \cite{DBLP:conf/iccv/BoutrosGKD23}*           & C-WF & 10K/50  & 98.98 & 89.30 & 90.97 & 91.25 & 86.87 & 91.47 & 23.44 & 69.69\\ 
      &     \textbf{NegFaceDiff (Ours)}     & C-WF & 10K/50  & \textbf{98.98} &	90.02 &	\textbf{91.67} &	\textbf{91.65}	& \textbf{88.82} &	\textbf{92.23} & \textbf{77.38}	& \textbf{86.11} 
\\ 
\hline
\end{tabular}%
}
\caption{Comparison of the FR model trained on the proposed NegFaceDiff dataset with SOTA on small and large-scale datasets. The presented results are verification accuracies in \%. Note that some of these results are not directly comparable 
due to the usage of different datasets to train each generative model, e.g. FFHQ \cite{FFHQ} (70k images) vs. C-WF \cite{DBLP:journals/corr/YiLLL14a} (500k images) vs. WebFace4M \cite{DBLP:conf/cvpr/ZhuHDY0CZYLD021} (4m images). Under the same settings, for example, DGM trained on C-WF \cite{DBLP:journals/corr/YiLLL14a}, NegFaceDiff outperformed IDiff-Face \cite{DBLP:conf/iccv/BoutrosGKD23} and other competitors on the considered evaluation benchmarks. 
The results marked with a star (*) are reproduced in this work as there was no publicly released FR model. Some of the previous works did not release their pretrained model and their generated data and there were no reported results on the large-scale benchmark IJB-C \cite{DBLP:conf/icb/MazeADKMO0NACG18}. These results are marked with (-). The first row presents the results achieved by authentic C-WF \cite{DBLP:journals/corr/YiLLL14a}, which are provided as a reference. 
}
\label{tab:sota}
\vspace{-3mm}
\end{table*}

\subsection{Comparison with SOTA Synthetic-based FR}
\label{sec:res_sota}



\paragraph{Face Recognition Evaluation}
Table \ref{tab:sota} presents the evaluation results of our proposed NegFaceDiff and previous SOTA methods on the small 
and large-scale 
evaluation benchmarks detailed in Section \ref{sec:eval_benchmarks}. NegFaceDiff and diffusion-based approaches outperformed digital rendering (DigiFace-1M \cite{DigiFace1M}) and GAN-based on all considered benchmarks with a clear margin. When sampling with a DM trained on FFHQ, NegFaceDiff presented competitive performance on small-scale benchmarks while significantly surpassing the baseline (IDiff-Face \cite{DBLP:conf/iccv/BoutrosGKD23}) on the large-scale and challenging benchmark IJB-C, where the TAR@FAR of 1e-5 and 1e-4 increased from 20.60\% and 62.60\% to 58.09\% and 73.93\%, respectively. When sampling with a DM trained on C-WF, NegFaceDiff increased the average baseline performance on small-scale benchmarks from 91.47\% to 92.23\%. This performance difference is more accentuated on IJB-C, with the TAR@FAR of 1e-5 and 1e-4 increasing from 23.44\% and 69.69\% to 77.38\% and 86.11\%, respectively. Note that the DMs' training datasets (FFHQ vs. C-WF) impact the synthetic data generated by these models, and, thus, the achieved FR performance.

\begin{table}[]
\resizebox{\linewidth}{!}{%
\begin{tabular}{ccccc|c||cc}
\hline
Models (DGMs Train Dataset)  & Indian & Caucasian & Asian & African & Avg. & STD & SER \\ \hline
C-WF \cite{DBLP:journals/corr/YiLLL14a} (-) & 88.50 & 93.38 & 85.38 & 86.62 & 88.47 & 3.52 & 2.21 \\ \hline
DigiFace1M \cite{DigiFace1M} (-) & 70.33  &  72.18    & 69.97 & 65.55   &  69.51  &  2.81   &  \textbf{1.23} \\ \hline
SFace \cite{Boutros2022SFace} (C-WF) & 67.07  & 73.15     & 68.07 & 63.57   &  67.97  &   3.96  &   1.36  \\
SFace2 \cite{10454585} (C-WF) & 71.48  & 77.63     & 71.55 & 67.23   &  71.97 &   4.28  &   1.46  \\
IDNet \cite{DBLP:conf/cvpr/KolfREBKD23} (C-WF) & 71.93  & 76.17     & 70.85 & 64.07   &  70.76  &  5.01   & 1.51  \\
ExFaceGAN \cite{ExFaceGAN} (FFHQ) & 71.58  & 73.80     & 71.07 & 62.20   &  69.66  &  5.11   & 1.44  \\ \hline
IDiff-Face \cite{DBLP:conf/iccv/BoutrosGKD23} (FFHQ) & 81.50  & 85.37     & 79.77 & 75.78   &   80.61  &   3.98  &  1.66  \\ 
\textbf{NegFaceDiff (Ours) (FFHQ)}    & 82.28  & 85.25  & 80.82 & 77.13   & 81.37     &   3.37  &  1.55  \\
DCFace \cite{DBLP:conf/cvpr/Kim00023} (C-WF) & 81.63  & 86.23     & 79.05 & 74.05   &  80.24 &  5.08   &  1.88  \\ 
IDiff-Face \cite{DBLP:conf/iccv/BoutrosGKD23} (C-WF) & 84.80 &	88.45 & 82.07 &	80.85 & 83.79 & 3.37 & 1.66 \\ 
\textbf{NegFaceDiff (Ours) (C-WF)}    &  \textbf{86.38} & \textbf{89.40} & \textbf{84.23} & \textbf{83.20} & \textbf{85.61} & \textbf{2.74} & 1.58
\end{tabular}
}
\vspace{-2mm}
\caption{Bias assessment of FR models trained on  NegFaceDiff dataset and SOTA. The ethnicity-specific results correspond to verification accuracies in \% on each ethnicity subset of RFW \cite{RFW}. STD and SER are bias metrics that can be calculated based on these verification accuracies. The best results for each metric (excluding authentic data) are highlighted in bold.}
\label{tab:rfw}
\vspace{-3mm}
\end{table}

\paragraph{Racial Bias Assessment}
\label{sec:rfw}
Table \ref{tab:rfw} presents the evaluation results of NegFaceDiff and previous SOTA synthetic-based methods on the RFW dataset \cite{RFW}, as detailed in Section \ref{sec:eval_benchmarks}. The results are reported as verification accuracies on each subset of RFW \cite{RFW} (Indian, Caucasian, Asian and African) as well as the average accuracy on these subsets. Following \cite{RFW}, we additionally reported STD and SER to assess the racial bias. Note that a lower STD and a SER closer to one indicate less bias. 
Diffusion-based approaches, DCFace \cite{DBLP:conf/cvpr/Kim0L22}, IDiff-Face \cite{DBLP:conf/iccv/BoutrosGKD23} and our NegFaceDiff, outperformed digital rendering (DigiFace1M \cite{DigiFace1M}) and GAN-based approaches in terms of average verification accuracies. The best average accuracy, 85.61\%, was achieved by our NegFaceDiff (sampling from a DM pre-trained on C-WF), ahead of 80.24\% by DCFace \cite{DBLP:conf/cvpr/Kim0L22} and 83.79\% by IDiff-Face \cite{DBLP:conf/iccv/BoutrosGKD23}.
In terms of STD where a lower value indicates less bias, our NegFaceDiff achieved the lowest value among previous SOTA works (2.74). As for the SER, i.e., the highest error divided by the lowest error, digital rendering and GAN-based approaches reached lower values compared to NegFaceDiff and other diffusion-based approaches. However, as previously mentioned, digital rendering and GAN-based approaches achieved very low verification accuracies compared to diffusion-based approaches. Among diffusion-based approaches, NegFaceDiff achieved the lowest SER (1.55 and 1.58 when based on a DM pre-trained on FFHQ and C-WF, respectively). It should be noted that some of the recent works presented in Table \ref{tab:sota} did not release their data or FR models and did not report any racial bias evaluation. Additionally, the lack of clear descriptions of their implementation details prevented us from replicating them to evaluate them on the RFW dataset \cite{RFW}.

\section{Conclusion}
This work introduced NegFaceDiff, a novel sampling method for identity-conditioned DMs that integrates negative contexts to enhance inter-class separability in synthetic face generation. By explicitly guiding the model away from unwanted identity features while preserving intra-class consistency, NegFaceDiff improves identity separability, as evidenced by the higher FDR and reduced EER. Our extensive evaluations demonstrate that FR models trained on NegFaceDiff datasets outperform those trained on synthetic datasets generated without negative contexts, achieving superior verification performance across multiple benchmarks, including large-scale IJB-C. Furthermore, our approach exhibits reduced racial bias, suggesting that incorporating negative conditions enhances both fairness and accuracy in synthetic-based FR. These findings highlight the potential of dual-conditioning techniques in generative models, paving the way for more robust, privacy-preserving, and ethically compliant FR systems.

\paragraph{Acknowledgment} This research work has been funded by the German Federal Ministry of Education and Research and the Hessian Ministry of Higher Education, Research, Science and the Arts within their joint support of the National Research Center for Applied Cybersecurity ATHENE.

{
    \small
    \bibliographystyle{ieeenat_fullname}
    \bibliography{main}
}

\end{document}